\documentclass[10pt,twocolumn,letterpaper]{article}

\usepackage{cvpr}              

\usepackage{graphicx}
\usepackage{amsmath}
\usepackage{amssymb}
\usepackage{booktabs}
\usepackage{multirow}
\usepackage{adjustbox}
\usepackage{pifont}

\usepackage{enumerate}
\usepackage{algorithm}
\usepackage{algpseudocode}
\usepackage{amsfonts}

\algrenewcommand\alglinenumber[1]{\tiny #1:}

\usepackage[pagebackref,breaklinks,colorlinks]{hyperref}

\usepackage[capitalize]{cleveref}
\crefname{section}{Sec.}{Secs.}
\Crefname{section}{Section}{Sections}
\Crefname{table}{Table}{Tables}
\crefname{table}{Tab.}{Tabs.}

\def\cvprPaperID{36} 
\def\confName{CVPR}
\def\confYear{2022}

\begin{document}

\title{Self-supervised Video Representation Learning with Cascade Positive Retrieval}


\author{
\textbf{Cheng-En Wu$^1$\thanks{Work done as a NEC Labs intern in 2021.}, Farley Lai$^2$, Yu Hen Hu$^1$, Asim Kadav$^2$}\\
$^1$ Department of Electrical and Computer Engineering, University of Wisconsin-Madison, WI, USA\\
$^2$ NEC Laboratories America, Inc., San Jose, CA, USA\\
\texttt{\{cwu356, yhhu\}@wisc.edu}\\
\texttt{farleylai@icloud.com}\quad
\texttt{asimkadav@gmail.com}
}
\maketitle

\begin{abstract}
   Self-supervised video representation learning has been shown to effectively improve downstream tasks such as video retrieval and action recognition.
   In this paper, we present the Cascade Positive Retrieval (CPR) that successively mines positive examples w.r.t. the query for contrastive learning in a cascade of stages.
   Specifically, CPR exploits multiple views of a query example in different modalities, where an alternative view may help find another positive example dissimilar in the query view. 
   We explore the effects of possible CPR configurations in ablations including the number of mining stages, the top similar example selection ratio in each stage, and progressive training with an incremental number of the final Top-k selection.
   The overall mining quality is measured to reflect the recall across training set classes.
   CPR reaches a median class mining recall of \textbf{83.3\%}, outperforming previous work by \textbf{5.5\%}.
   Implementation-wise, CPR is complementary to pretext tasks and can be easily applied to previous work. 
   In the evaluation of pretraining on UCF101, CPR consistently improves existing work and even achieves state-of-the-art R@1 of \textbf{56.7\%} and \textbf{24.4\%} in video retrieval as well as \textbf{83.8\%} and \textbf{54.8\%} in action recognition on UCF101 and HMDB51.
The code is available at \url{https://github.com/necla-ml/CPR}.

\end{abstract}



\section{Introduction}
\label{sec:intro}



Recently, large-scale self-supervised pretraining such as BERT\cite{Devlin2019-gx} and DINO~\cite{Caron2021-hr} has been shown to improve the representations and potentially outperform its supervised counterpart.
Most approaches revolve around proposing pretext tasks \cite{Xie2018-fr, Kim2019-os, Xu2019-uv, Han2019-au, Benaim2020-pq, Jenni2020-wc, Luo2020-gu, Yao2020-gv, Wang2020-hs} based on instance discrimination to learn  representations by matching or classifying specific relationships between the query example and its augmented variants with the objective to minimize the contrastive loss \cite{Van_den_Oord2018-hk} and other predictive losses. 
However, few address the lack of true positives (TP) other than the query example variants and likely harmful false negatives uniformly sampled from the entire dataset \cite{Chuang2020-ey}.
Previous work CoCLR~\cite{Han2020-od} demonstrates the significant performance gap with the upper bound achieved in a supervised contrastive setting using the labels for TP as in \cite{Khosla2020-vt}.

We are inspired by related work~\cite{Tian2020-hz, Wang2019-qy, Han2020-jb, Huang2021-ds, Wang2021-id, Rai2021-hr} that exploits multi-views of video to learn the representations through the correspondences between different modalities.
Previous work \cite{Suh2019-dh, Jin2018-cn, Han2020-od} incorporating hard example mining in metric learning, object detection and action recognition further motivates the necessity of positive example mining in self-supervised representation learning.
As for video representation learning, hard positive examples in the RGB view may be mined from the motion view despite seemingly different background appearances.
On the other hand, hard positive examples in the motion view may be mined from the RGB view as the motions can differ significantly from various camera angles while the background remains similar in the RGB view for actions in the same class.
CoCLR~\cite{Han2020-od} shows mining in the alternative view during training improves the representations and downstream task performance.
Nonetheless, it is not necessarily sufficient for mining only once in a single view to prevent sampling false positives (FP).

To address this issue, we propose the Cascade Positive Retrieval (CPR) and systematically explore the design space of positive example mining.
The idea is to refine the mining successively in a cascade of stages across different views as search with filters to be applied progressively.
For instance, given a query example, one may first select those with similar background in the RGB view, then further filter out those dissimilar in the motion view and so on.
Apparently, the number of mining stages and the selection ratio in each stage matter.
The goal is to conclude the strategy for effective positive example mining and make it applicable to existing work.
Moreover, it remains unclear of the overall mining quality in terms of the recall across training set classes despite the R@1 mining retrieval recall by CoCLR~\cite{Han2020-od}.
We measure and compare the mining quality that suggests correlation with the resulting performance in ablations.

In short, we make the following contributions: 
\begin{enumerate}[1{.)}]
\item We propose the Cascade Positive Retrieval for self-supervised learning (SSL) of video representations that complements pretext tasks and can be applied to existing work easily regardless of the SSL framework used.
\item We apply CPR to previous work and observed consistent improvement in downstream video retrieval and action recognition. We then extensively explore the design space of mining configurations in ablations w.r.t. the number of stages in the cascade, the top similar example selection ratio in each stage and the progressive training regime.
\item We measure the mining quality of CPR in terms of the positive mining recall denoting each time the fraction of TPs in the final stage Top-$k$ selected as the positive set, and the class mining recall representing the fraction of distinct TPs selected from a class in one training epoch.

\item We evaluate the transfer performance in video retrieval and action recognition on UCF101 and HMDB51 from pretraining on UCF101 with CPR applied to an existing work, achieving state-of-the-art (SOTA) results.

\end{enumerate}

\section{Related Work}
\label{sec:r_work}



\noindent \textbf{Self-supervised Learning.}
Large-scale representation learning through self-supervision has achieved great success in multiple fields including natural language processing (NLP) and computer vision (CV). 
In NLP, the general idea is to build a language model that learns to predict masked out words as in BERT\cite{Devlin2019-gx}.
In CV, the feature extraction backbone is trained to learn representations based on instance discrimination that works on both images and videos.
The instance discrimination views an example and its augmented variants as positive while the other examples are treated as negative.
A typical objective is to minimize the contrastive loss that encourages positive examples to be similar in representations while pushing away negative examples.
Many SSL frameworks were proposed in recent years such as SimCLR\cite{Chen2020-dm}, BYOL\cite{Grill2020-zb}, MoCo\cite{He2019-wo, Chen2020-go} and SwAV\cite{Caron2020-uu} to facilitate systematic composition of numerous pretext tasks that augment the input examples and formulate the contrastive loss, delivering competitive performance in comparison with supervised counterparts.
In this paper, we focus on improving self-supervised video representation learning from the perspective of hard positive example mining and show our method can be easily applied to existing work regardless of a particular SSL framework used or not.


\noindent \textbf{Video Representation Learning.}
In contrast with SSL of images, videos enables rich spatiotemporal augmentation to generate diverse positive and negative example clips from sampled frames.
Common pretext tasks include future prediction\cite{Han2020-jb} and speed prediction\cite{Benaim2020-pq, Jenni2020-wc, Wang2020-hs, Yao2020-gv} to infer the relationship between clips and the pace a clip is sampled.
Other tasks may require to sort out the ordering of frames or clips\cite{Xu2019-uv}, solve jigsaw puzzles\cite{Kim2019-os}, match features in different modalities\cite{Wang2019-qy, Wang2021-id, Huang2021-ds, Rai2021-hr} or group visual entities based on co-occurrences in space and time\cite{Isola2016-yy}.
We target the video domain as videos in multiple views potentially provide opportunities to mine hard positive examples in the query class.
Nonetheless, the proposed method is not limited to video tasks or specific pretext tasks.
Instead, we aim to complement existing approaches with hard positive example mining.



\noindent \textbf{Hard Example Mining.} Hard example mining in supervised learning is well studied in metric learning and other CV tasks.
In metric learning, the goal is the push away those hard negative examples but the challenge is the intractable computational overhead over large datasets as the embedding is updated constantly.
One possible solution is to efficiently sample negative instances in nearest classes as in deep metric learning \cite{Suh2019-dh}.
Regarding positive example mining, InvP~\cite{wang2020unsupervised} selects positive examples that preserve high semantic consistency through a recursive k-nearest neighbors graph. 
In addition, CMA~\cite{morgado2021audio} introduces the cross-modal agreement that discovers positive examples highly similar in both audio and visual feature space through multi-view learning~\cite{Tian2020-hz}.

In video object detection, \cite{Jin2018-cn} leverages the temporal consistency to identify hard negative and positive examples from detection misses and isolated detection in consecutive video frames.

In the case of SSL, it is challenging for no labels and the representation learning is limited to the augmentations of the query example with instance discrimination for the lack of hard positive examples in the query class.
Worse, the negative examples are uniformly sampled and potentially include false negatives (FN).
This is called the sampling bias in \cite{Chuang2020-ey} and a possible solution is to reweight the positive and negative terms in the contrastive loss for correction given the estimated class priors~\cite{Chuang2020-ey, Robinson2021-hv}.

On the other hand, as with the video object detection, self-supervised video representation learning may exploit multi-views of video clips to mine hard positive examples.
CoCLR~\cite{Han2020-od} mines positive examples from action recognition datasets given a query example in the RGB view with its corresponding motion or flow view.
Intuitively, this may help find positive examples with similar motions despite dissimilar background and vice versa.
Our work further explores the possibilities to mine diverse positive examples in the query example class as CoCLR only mines positive examples in one view at a time.
Chances are out of those with similar motions, top instances similar in the RGB view could be more likely the true positives.
Therefore, we reshape the positive example mining as a cascade refining process between different video views.
While CoCLR measures R@1 for mining retrieval recall, we further evaluate the mining quality in terms of the overall mining recall across the classes throughout training in reflection of the coverage of distinct class instances.
The metric is expected to correlate with the resulting performance w.r.t. the upper bound in the supervised contrastive setting where the mining recall is essentially perfect for all the class instances being selected during training.


\section{Proposed Method}
\label{sec:method}


\begin{figure*}[]
  \begin{center}
    \includegraphics[width=0.80\textwidth]{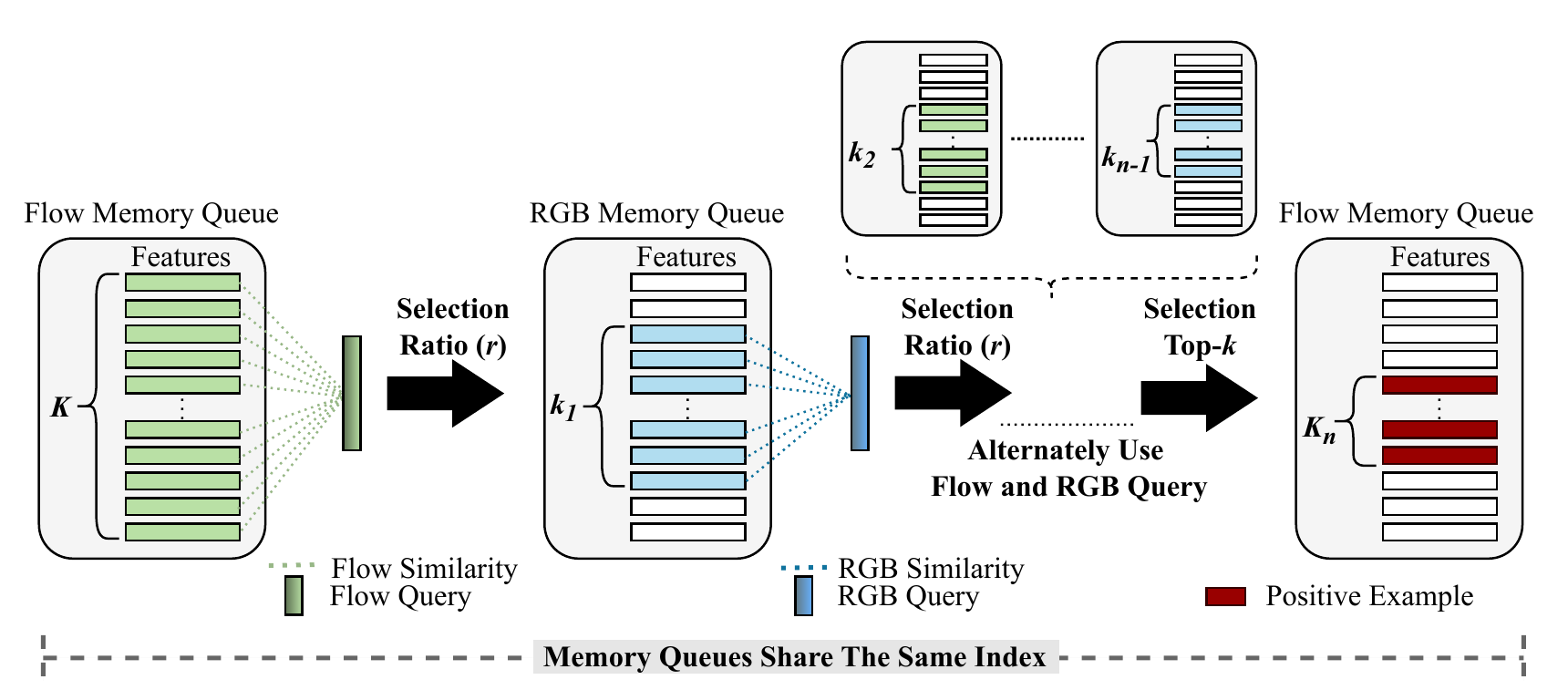}
  \end{center}
  \caption{Overview of CPR in mining alternately from both RGB and flow views for one cascade retrieval.
  $K$ denotes the size of the memory queues storing instance features in both views.
  Given a query example in the RGB view, the mining starts with selecting top $k_1$ most similar instances from the flow memory queue at stage 1, top $k_2$ most similar instances from the RGB memory queue at stage 2, and so on up to top $k_{n-1}$ from the RGB memory queue at stage $n$-$1$, where $k_1$, $k_2$, ..., $k_{n-1}$ are values derived from the number of instances selected from the previous stage multiplied by a fixed selection ratio (e.g. 0.5) at each stage.
  Unlike previous stages, the Top-$k$ most similar instances $k_n$ at the final stage are selected to form the positive set.}
  \label{fig:framework}
\end{figure*}


\begin{center}
\scalebox{0.9}{
\begin{minipage}{1.0\columnwidth}
\begin{algorithm}[H]
\caption{CPR: Cascade Positive Retrieval}\label{alg:CPR}
\scriptsize
\begin{algorithmic}[1]
\Require $MB, C, S, B, V, r, v_q, q_v, q, q^+$
\Ensure $E(c), K(c, s), SV(s), B(e)$
\Ensure $select(f_v, candidates_v, r), topk(f_v, candidates_v, k)$

\State $C$                                              \Comment{range of training cycles}
\State $S \gets 1..n$                                   \Comment{range of CPR stages}
\State $E(c) \in \mathbb{Z}^+$                          \Comment{epochs given a training cycle}
\State $K(c, s) \in \mathbb{Z}^+$                       \Comment{Top-$k$ to select at stage s in cycle c}
\State $r \in \mathbb{R}^+$                             \Comment{selection ratio before the last stage}
\State $select(f_v, candidates_v, r) \in \mathbb{Z}^+$  \Comment{select top similar instances by ratio}
\State $topk(f_v, candidates_v, k)  \in \mathbb{Z}^+$   \Comment{select top k similar instances}
\State $SV(s) \in \mathbb{Z}^+$                         \Comment{given a view at stage s}
\State $V \in \{v_1, v_2, ...\}$                        \Comment{set of views}


\For {$c \in C$}
    \For {$e \in E(c)$}
        \For {$(q, q+) \in B(e)$}
            \For {$v \in V$}
                \If{$v == v_q$}
                    \State $f_{q_{v_q}} \gets encoder_{v_q}(q_{v_q})$
                    \State $f_{q^+_{v_q}} \gets encoder^{ema}_{{v_q}}(q^+_{v_q})$
                \Else
                    \State $f_{q^+_v} \gets encoder^{fixed}_{{v}}(q^+_v)$
                \EndIf
            \EndFor
            \For {$s \in S$}
                \State $v \gets SV(s)$
                \If {$s == 1$}
                    \State $pos \gets select(f_{q^+_v}, MB_v, r)$
                \ElsIf {$s == n$}
                    \State $pos \gets topk(f_{q^+_v}, pos_v, K(c, s))$
                    \State $pos = \{q^+, pos\}$
                    \State $neg = MB \setminus pos$
                \Else
                    \State $pos \gets select(f_{q^+_v}, pos_v, r)$
                \EndIf
            \EndFor
            \State $loss \gets MIL\_NCE(q_{v_q}, pos_{v_q}, neg_{v_q})$
			\State $optimize(encoder_{v_q}, loss)$
			\State $update(MB, f_q^+)$
        \EndFor
    \EndFor
\EndFor
\end{algorithmic}
\end{algorithm}
\end{minipage}
}
\end{center}

In this section, we first revisit the concept of contrastive learning with different discrimination learning objectives. 
Next, we present CPR in Algorithm~\ref{alg:CPR}, detailing the cascade positive retrieval for mining examples in general.

\subsection{Instance Discrimination}
\label{method:nce_loss}

Self-supervised video representation learning based instance discrimination where each instance serves as its own class has been shown effective with the contrastive loss of InfoNCE\cite{Van_den_Oord2018-hk}.
Specifically, given a set of videos $V$, a video clip $v_i$ is a number of frames sampled from a video in $V$ and its positive variant $v_{i}^{+}$ that can be an augmentation or another clip sampled from the same video, forming a positive pair ($v_i$, $v_i^+$).
On the other hand, a set of negative examples $N^-$ consists of those clips $v_{j}^{-}$, $j \neq i$. 
These clips are fed into a query encoder and a key encoder to obtain the visual representations.
The output features of the query, its positive augmentation and negative keys are denoted by $q_i$, $q_i^+$, and $k_j^{-}$ respectively.
The InfoNCE loss is defined as follows:
\begin{equation}
\label{eq:LN}
\footnotesize
\mathcal{L}_{N}=-\log\frac{\exp(q_{i}^{} \cdot q_{i}^{+} /\tau)}{\exp(q_{i}^{} \cdot q_{i}^{+} /\tau)+\sum_{j=1}^{N}\exp(q_{i}^{} \cdot k_{j}^{-} /\tau)}
\end{equation}
where the similarity is measured by dot product with a temperature hyperparameterper $\tau$ to adjust its scale. 
Intuitively, InfoNCE encourages to pull positive pairs closer while pushing away negative pairs.

\subsection{Multi-instance Discrimination}
\label{method:multince_loss}
In the case of multiple positive pairs, Multi-Instance InfoNCE or MIL-NCE proposed in \cite{Miech2020-ea} is defined as follows: 
\begin{equation}
\label{eq:LM}
\footnotesize\mathcal{L}_{M}=-\log\frac{\sum_{p \in P}\exp(q_{i}^{} \cdot q_{p}^{+} /\tau)}{\sum_{p \in P}\exp(q_{i}^{} \cdot q_{p}^{+} /\tau)+\sum_{j=1}^{N}\exp(q_{i}^{} \cdot k^{-}_{j} /\tau)}
\end{equation}
where $P$ is a positive set containing positive augmentation of the query and other keys with the same label as the query. 
For example, in an action video dataset, a \textit{fencing} positive set includes the augmentation of the query video and other videos with the \textit{fencing} label. 

\subsection{Cascade Positive Retrieval}
\label{method:cascade}

\begin{figure*}[]
  \begin{center}
    \includegraphics[width=0.84\textwidth]{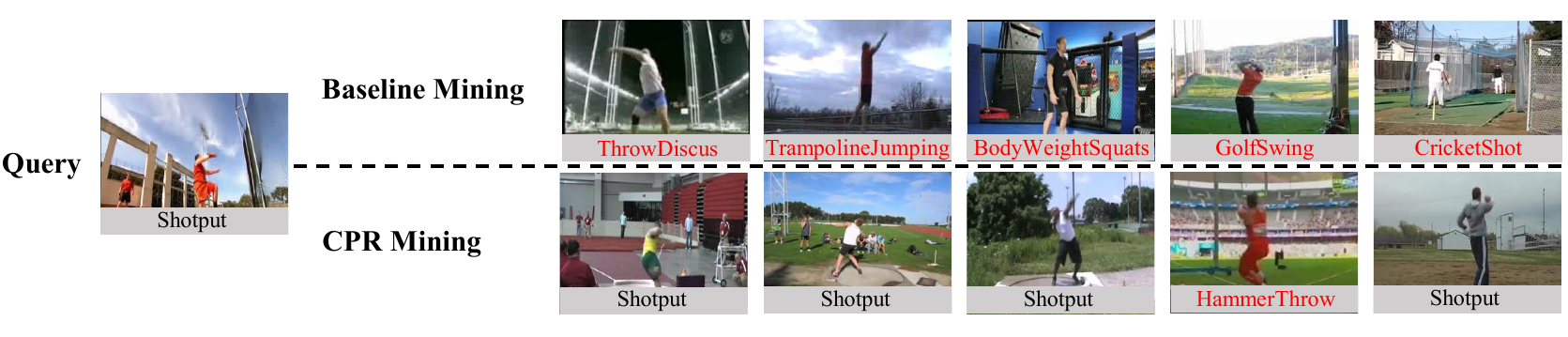}
  \end{center}
  \caption{Qualitative Top-$5$ mining comparison with wrong selection in red.}
  \label{fig:exp_pos_visual}
\end{figure*}

In view of issues with instance discrimination including the lack of other non-augmented positives and potential false negatives, previous work CoCLR~\cite{Han2020-od} has proposed to mine positive examples in an alternative view other than the query view.
However, there is a possibility that CoCLR suffers from FPs with similar motion patterns from the flow view because the mining in the alternative view is only done once such that some actions with very similar motion patterns such as \textit{Shouput} and \textit{ThrowDiscus} may be wrongly selected and confuse the model as shown in Figure~\ref{fig:exp_pos_visual}.
Unlike CoCLR mining heavily dependent on a single view, our CPR fully exploits the advantage of multi-views to improve the mining quality.
Figure~\ref{fig:framework} illustrates that in one cascade of positive retrieval, CPR alternates between the RGB and flow views to mine a top number of positive examples with most similar appearances and motions as the query clip.

When applying CPR to existing work, there are many possible configurations and hyperparameters to consider as described in Algorithm~\ref{alg:CPR} that assumes a memory bank $MB$ storing encoded instance features in different views, a progressive training schedule in cycles, the number of epochs in one cycle, the number of mining stages in one cascade, the selection ratio of top similar examples at each stage and etc.
Specifically, the algorithm iterates through each cycle $c$ and epoch $e$ to train with query examples in batches $B$.
Each batch consists of query examples and their positive variants from augmentation or sampling as $q$ and $q^+$.
In the beginning of the batch processing, the representation encoder to train in the query view,  $encoder_{v_q}$, encodes the query examples $q_{v_q}$ and produces the features $f_{q_{v_q}}$.
Those $q^+$ may be encoded in the query view with the momentum $encoder^{ema}_{{v_q}}(q^+_{v_q})$ and in the other views with frozen $encoder^{fixed}_{{v}}(q^+_v)$.
Next, CPR retrieves the most similar examples in successive stages $S$ given a selection ratio $r$ used at stages before the last one and a Top-$k$ for the final stage selection determined by the current cycle $c$ and stage $s$.
Note that the mining always uses the features of positive query variants to measure the similarities with those stored in $MB$ by view.
Eventually, a set of Top-$k$ most likely positives are selected at the last stage as $pos$ and combined with $q^+$.
The other instances in $MB$ are viewed as negatives $neg$.
Then the MIL-NCE loss is computed given the query examples, mined positives and negatives to optimize the encoder in the query view.
Afterwards, the memory bank $MB$ is updated with the newly encoded query example variants for the next batch training iteration.
In the next section, we will evaluate the effects of changing CPR hyperparameters in ablations as well as compare the performance with SOTAs.



\section{Experiments}
\label{sec:exp}

\subsection{Setup}
\label{sec:exp_setup}
\noindent \textbf{Dataset.} In this section, we conduct ablation studies and evaluate CPR on two action video datasets:\\
\indent \textbf{UCF101}~\cite{Soomro2012-ge} contains 13K videos in 101 human action classes at more diverse camera angles than HMDB51. Out of the three splits of the dataset, the first one is used for our ablations, pretraining, and downstream task evaluations.
\\
\indent \textbf{HMDB51}~\cite{Kuehne2011-wa} consists of 7K videos in 51 human action categories. 
The dataset is divided into three splits. We use the first split to conduct two downstream tasks in video retrieval and action recognition.
 

\noindent \textbf{Implementations.} We apply CPR to previous work IIC~\cite{Tao2020-tb} and CoCLR~\cite{Han2020-od}.
While the latter uses MoCo\cite{He2019-wo}, CPR is not dependent on specific SSL frameworks.
For fair comparison, we use exactly the same hyperparameters as previous work and only plug in CPR to construct the positive and negative sets for computing the MIL-NCE loss.
If necessary, we even retrain previous work for the same number of epochs to compare with the reproduced results.
More details can be found in the supplemental materials.

\noindent \textbf{Data Preprocessing}: The data preparation follows previous work respectively.
As for CoCLR~\cite{Han2020-od}, a clip in both RGB and flow views is randomly sampled from 32 consecutive frames in the video.
Each frame is randomly cropped and resized to 128×128 pixels.
We apply the same data augmentations including horizontal flips, color jittering and Gaussian blur to the clips. 
Note that Gaussian blur is not used for downstream tasks.
To generate optical flow maps from the video, we use TV-L1~\cite{Zach2007-bq} to extract the flow view with a third channel filled with zeros.
The features are clipped in the range of 20 pixels and rescaled from $[-20, 20]$ to $[0, 255]$.
In contrast, the motion view for IIC~\cite{Tao2020-tb} is based on frame difference residuals.

\noindent \textbf{Self-supervised Pretraining on UCF101.} 
For IIC~\cite{Tao2020-tb}, we train from scratch with CPR under NPID~\cite{wu2018unsupervised}. 
For CoCLR~\cite{Han2020-od}, we begin with the released RGB and flow models pretrained with InfoNCE as there is no positive mining in the initialization. 
Next at the co-training stage, the RGB and flow models are alternately trained for 400 epochs on 2 GPUs, each with a batch size of 16.
That is the same number of epochs as two cycles in CoCLR.

\noindent \textbf{Video Retrieval.} We evaluate video retrieval as a downstream task on both UCF101 and HMDB51 based on extracted features from the pretrained model without finetuning. Following the test protocol in~\cite{Luo2020-gu,Xu2019-uv}, we take a video in the test set as a query and use it to retrieve $k$-nearest neighbors in its corresponding training set. 
The recall at $k$ (R@$k$) serves as the evaluation metric, which means if one of the retrieved top $k$ nearest neighbors is from the same class as the query, it is counted as a correct retrieval result.

\noindent \textbf{Action Recognition.} In addition to video retrieval, we also evaluate the action recognition performance of the pretrained models on UCF101 and HMDB51.
The pretrained models are transferred as the feature extraction backbone for downstream tasks. 
Two scenarios including \textbf{linear probing} and \textbf{finetuning} are considered respectively. 
For linear probing, we freeze the backbone while training the linear classifier only. 
For finetuning, we train the entire network including the backbone and the linear classifier. 
The training and evaluation protocols essentially follow previous work for fair comparison even with test time augmentation used.

\begin{table}[]
\centering
\begin{tabular}{l|ccccc}
\hline
Stages ($s$) & R@1 & R@5 & R@10 & Probe & Finetune \\ \hline
$s=1$ & 45.1 & 64.0 & 71.9 & 60.0 & 69.5 \\
$s=3$ & 46.5 & 64.5 & 72.0 & 60.0 & 69.6 \\
$s=5$ & 47.5 & 65.1 & 73.3 & 60.2 & 70.6 \\
$s=7$ & 47.8 & 66.2 & 74.6 & 60.4 & 71.3 \\ \hline
\end{tabular}
\caption{Ablations with CPR applied to CoCLR w.r.t. the number of stages. CoCLR is a special case with CPR in only one stage as $s=1$ where only the Top-$5$ positive candidates are selected.}
\label{table:stage}
\end{table}

\begin{table}[]
\centering
\begin{tabular}{l|ccccc}
\hline
SR ($r$) & R@1 & R@5 & R@10 & Probe & Finetune \\ \hline
$r=0.8_{(s=3)}$ & 46.1 & 63.9 & 72.4 & 60.0 & 69.8 \\
$r=0.5_{(s=3)}$ & 46.5 & 64.5 & 72.0 & 60.0 & 69.6 \\ \hline
$r=0.8_{(s=7)}$ & 46.2 & 63.6 & 71.7 & 59.6 & 69.9 \\
$r=0.5_{(s=7)}$ & 47.8 & 66.2 & 74.6 & 60.4 & 71.3 \\ \hline
\end{tabular}
\caption{Results for CPR applied to CoCLR with varied selection ratios but a fixed number of stages $s$.}
\label{table:selection}
\end{table}

\begin{table}[]
\centering
\begin{tabular}{l|cccc}
\hline
IIC(+CPR) & R@1 & R@5 & R@10 & R@20 \\ \hline
Baseline-Top-$5$ & {36.2} & {53.5} & {63.6} & {72.7} \\
Prog-Top-$1$ & 39.4 & 57.5 & 67.6 & 77.2 \\
Prog-Top-$2$ & 42.5 & 60.5 & 69.0 & 77.5 \\
Prog-Top-$3$ & 44.1 & 62.3 & 70.2 & 77.9 \\
Prog-Top-$4$ & 45.3 & 63.2 & 70.4 & 78.2 \\
Prog-Top-$5$ & 46.2 & 63.6 & 71.4 & 79.2 \\ \hline
\end{tabular}
\caption{Improvements in video retrieval with progressive training when CPR is applied to IIC~\cite{Tao2020-tb} in 5 cycles with incremental Top-$k$ selection. 
The baseline is trained with the same number of total epochs in the 5 cycles with a fixed Top-$k$ selection at the last stage.}
\label{table:prog}
\end{table}

\begin{table}[]
\centering
\begin{tabular}{l|ccc}
\hline
Settings & PMR & R@1 & Finetune \\ \hline
$s=1$ (CoCLR) & 35.1 & 45.1 & 69.5 \\
$s=3, r=0.5$ & 35.8 & 46.5 & 69.6 \\
$s=5, r=0.5$ & 37.4 & 47.5 & 70.6 \\
$s=7, r=0.5$ & 38.9 & 47.8 & 71.3 \\ \hline
\end{tabular}
\caption{Mean PMR and R@1 measured in the last epoch as well as fine-tuning results w.r.t. different CPR configurations.}
\label{table:capability}
\end{table}

\begin{figure}[h]
  \begin{center}
    \includegraphics[width=0.84\columnwidth]{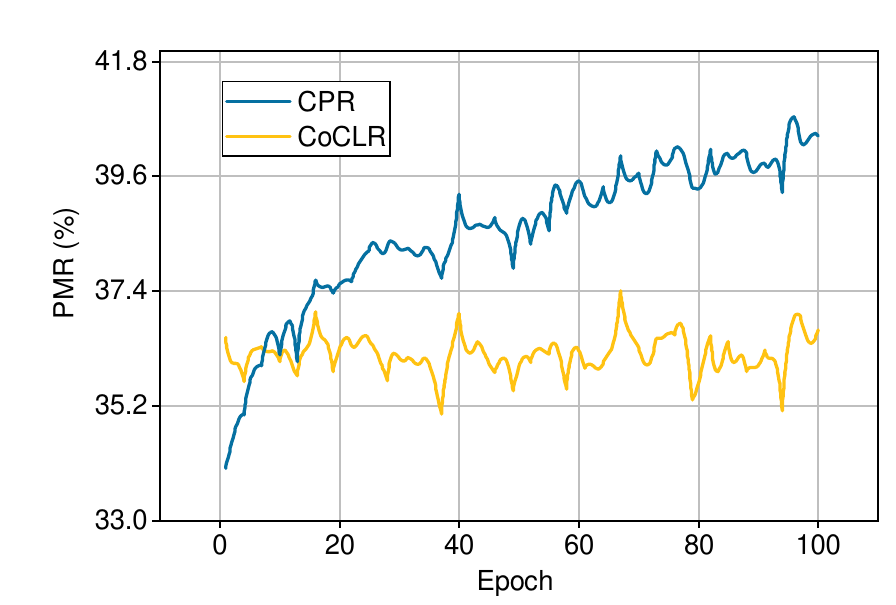}
  \end{center}
  \caption{Positive Mining Recall (PMR) for 100 training epochs. The results are generated by both models pretrained on UCF101.}
  \label{fig:full_PMR}
\end{figure}

\begin{figure}[h]
  \begin{center}
    \includegraphics[width=0.84\columnwidth]{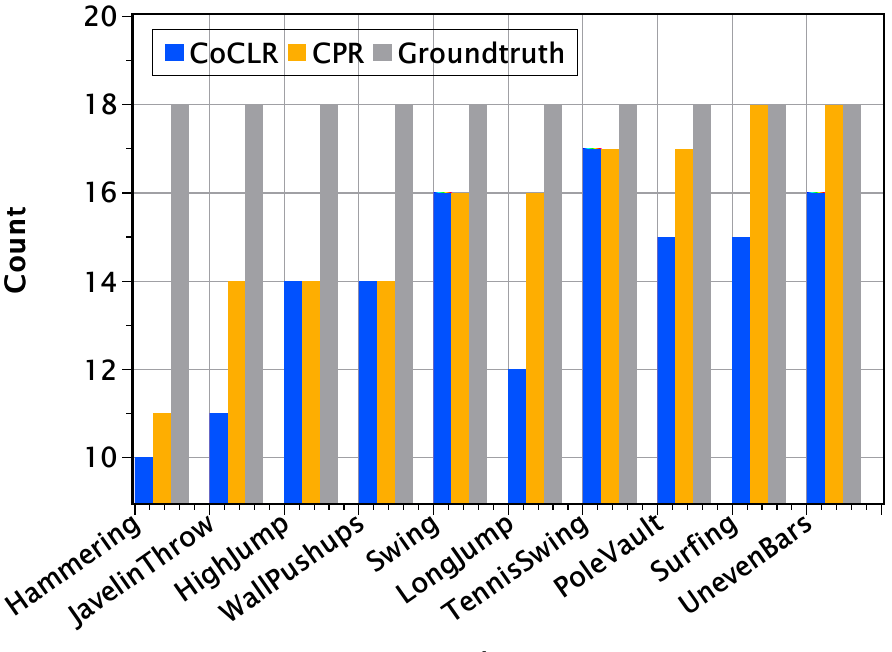}
  \end{center}
  \caption{TP class instance mining counts in selected classes.}
  \label{fig:bar_pos}
\end{figure}

\begin{figure}[h]
  \begin{center}
    \includegraphics[width=0.64\columnwidth]{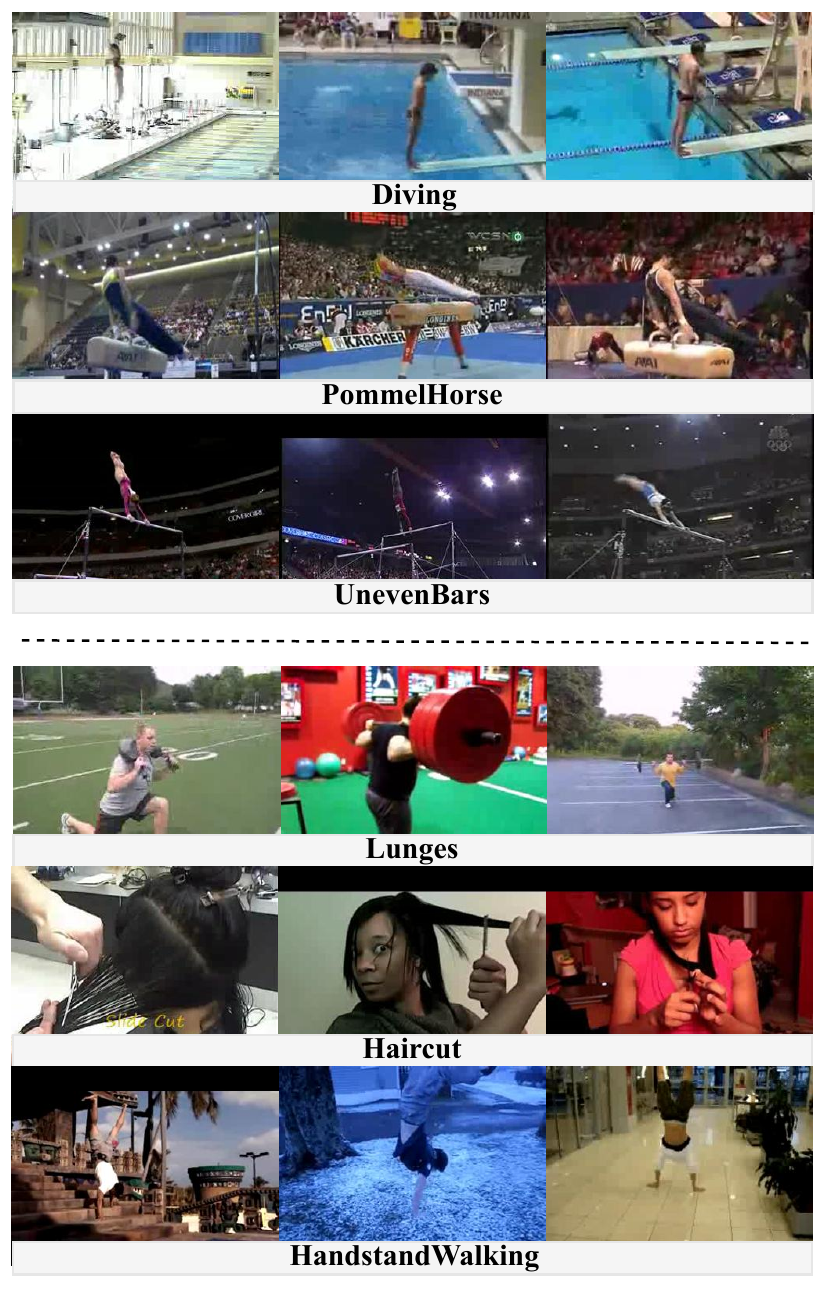}
  \end{center}
  \caption{Visualization of the action classes that are ranked in CMR. The top half of the figure shows the Top-3 classes while the bottom of the figure shows the Bottom-3 classes.}
  \label{fig:top3_bottom3}
\end{figure}

\begin{table}[h]
\begin{tabular}{l|cccc}
\hline
Method & \multicolumn{1}{l}{R@1} & \multicolumn{1}{l}{R@5} & \multicolumn{1}{l}{R@10} & \multicolumn{1}{l}{Finetune} \\ \hline
IIC~\cite{Tao2020-tb}  & 34.8 & 51.6 & 60.8 & 71.8 \\
IIC(+CPR) & 46.2 & 63.6 & 71.4 & 73.1 \\ \hline
CoCLR & 45.1 & 64.1 & 71.9 & 69.5 \\
CoCLR(+CPR) & 47.8 & 66.2 & 74.6 & 71.3 \\
UberNCE & 70.3 & 81.7 & 86.8 & 80.7 \\ \hline
\end{tabular}
\caption{Summary of improvements over IIC and reproduced CoCLR~\cite{Han2020-od} with CPR on UCF101. UberNCE is reproduced in the supervised contrastive setting serving as the upper bound.}
\label{table:trans}
\end{table}


\begin{table*}[]
\centering
\begin{tabular}{lcc|rrrr|rrrr}
\hline
\multirow{2}{*}{Method} & \multicolumn{1}{l}{\multirow{2}{*}{Year}} & \multicolumn{1}{l|}{\multirow{2}{*}{Backbone}} & \multicolumn{4}{c|}{UCF101} & \multicolumn{4}{c}{HMDB51} \\ \cline{4-11} 
 & \multicolumn{1}{l}{} & \multicolumn{1}{l|}{} & \multicolumn{1}{l}{R@1} & \multicolumn{1}{l}{R@5} & \multicolumn{1}{l}{R@10} & \multicolumn{1}{l|}{R@20} & \multicolumn{1}{l}{R@1} & \multicolumn{1}{l}{R@5} & \multicolumn{1}{l}{R@10} & \multicolumn{1}{l}{R@20} \\ \hline
VCOP~\cite{Xu2019-uv} & 2019 & R(2+1)D & 14.1 & 30.3 & 40.4 & 51.1 & 7.6 & 22.9 & 34.4 & 48.8 \\
VCP~\cite{Luo2020-gu} & 2020 & R3D-50 & 18.6 & 33.6 & 42.5 & 53.5 & 7.6 & 24.4 & 36.3 & 53.6 \\
MemDPC-RGB~\cite{Han2020-jb} & 2020 & R-2D3D & 20.2 & 40.4 & 52.4 & 64.7 & 7.7 & 25.7 & 40.6 & 57.7 \\
MemDPC-Flow~\cite{Han2020-jb} & 2020 & R-2D3D & 40.2 & 63.2 & 71.9 & 78.6 & 15.6 & 37.6 & 52.0 & 65.3 \\
IIC~\cite{Tao2020-tb} & 2020 & R3D-18 & 42.4 & 60.9 & 69.2 & 77.1 & 19.7 & 42.9 & 57.1 & 70.6 \\
PacePred~\cite{Wang2020-hs} & 2020 & R3D-18 & 23.8 & 38.1 & 46.4 & 56.6 & 9.6 & 26.9 & 41.1 & 56.1 \\
CoCLR-RGB~\cite{Han2020-od} & 2020 & S3D & 53.3 & 69.4 & 76.6 & 82.0 & 23.2 & 43.2 & 53.5 & 65.5 \\
CoCLR-Flow~\cite{Han2020-od} & 2020 & S3D & 51.9 & 68.5 & 75.0 & 80.8 & 23.9 & 47.3 & 58.3 & 69.3 \\
DSM~\cite{Wang2021-id} & 2021 & I3D & 17.4 & 35.2 & 45.3 & 57.8 & 7.6 & 23.3 & 36.5 & 52.5 \\
STS~\cite{Wang2021-na} & 2021 & R3D-18 & 38.3 & 59.9 & 68.9 & 77.2 & 18.0 & 37.2 & 50.7 & 64.8 \\
CMD~\cite{Huang2021-ds} & 2021 & C3D & 41.7 & 57.4 & 66.9 & 76.1 & 16.8 & 37.2 & 50.0 & 64.3 \\
VCLR~\cite{Kuang2021-og} & 2021 & R2D-50 & 46.8 & 61.8 & 70.4 & 79.0 & 17.6 & 38.6 & 51.1 & 67.6 \\
MFO~\cite{Qian2021-lm} & 2021 & R3D-18 & 39.6 & 57.6 & 69.2 & 78.0 & 18.8 & 39.2 & 51.0 & 63.7 \\
MCN~\cite{Lin2021-md} & 2021 & R3D-18 & 53.8 & 70.2 & 78.3 & 83.4 & 24.1 & 46.8 & 59.7 & 74.2 \\ \hline
CoCLR-RGB(+CPR) &  & S3D & 50.4 & 66.1 & 73.0 & 80.4 & 18.2 & 40.1 & 52.5 & 66.7 \\
CoCLR-Flow(+CPR) &  & S3D & \textbf{56.7} & \textbf{75.5} & \textbf{82.2}  & \textbf{88.2} & \textbf{24.4} & \textbf{48.5} & \textbf{62.4} & \textbf{74.3} \\ \hline

\end{tabular}
\caption{Comparison with SOTA video retrieval on UCF101 and HMDB51. Note that all methods are pretrained on UCF101.}
\label{table:retrieval}
\end{table*}

\subsection{Ablation Study}
\label{sec:ablation}
In this section, we explore CPR in numerous configurations. 
All experiments are conducted on UCF101 following the setup mentioned in Section~\ref{sec:exp_setup} except for the number of training epochs fixed at 100 for pretraining and finetuning respectively.
Unless said otherwise, the ablations are based on application of CPR to CoCLR.

\noindent \textbf{Number of Stages.} Our CPR mines positive examples in a cascade of multiple stages. 
It is necessary to demonstrate the influence of this hyperparameter given a fixed selection ratio $0.5$ for positive selection across stages before the last one and Top-$5$ for the last stage.
As shown in Table~\ref{table:stage}, with more mining stages, the model may learn better representations for the downstream task and the best performance is achieved in the configuration with 7 stages.
As a result, we use 7 stages in later evaluation with other SOTAs.

\noindent \textbf{Selection Ratio.} In this ablation, we set the selection ratio (SR) to 0.5 and 0.8 respectively to evaluate the impact of SR before the final stage that uses fixed Top-$5$. 
The total number of stages are set to 5 and 7 for comparison.
It is observed that no matter in 5 or 7 stages, a smaller SR can get better performance in Table~\ref{table:selection}.
Hence, we choose $SR=0.5$ when comparing with other SOTAs.

\noindent \textbf{Progressive Training.} This configuration examines the training regime of the Top-$k$ selection at the last stage.
Specifically, is it better to train with a fixed Top-$k$ or the training should be progressive with an incremental number of Top-$k$.
The conclusion is likely model architecture dependent as we see the improvement with IIC shown in Table~\ref{table:prog} in terms of video retrieval recalls but little with CoCLR.
Therefore, progressive training will not be applied to CoCLR in other evaluations.

\noindent \textbf{CPR Mining Quality.} While CoCLR~\cite{Han2020-od} measures R@1 for the mining retrieval against the ground truth (GT) throughout training, it remains unclear how many TPs are actually mined each time and throughout the training.
Therefore, we measure the positive mining recall and class mining recall defined in Eq.~\ref{eq:pos_recall} and Eq.~\ref{eq:cls_recall}:
\begin{equation}
\label{eq:pos_recall}
\footnotesize
Positive~Mining~Recall=\frac{\#~TP}{Size~of~Positive~Set}
\end{equation}
\begin{equation}
\label{eq:cls_recall}
\footnotesize
Class~Mining~Recall=\frac{\#Distinct~TP~Selected}{\#Total~Class~Instances}
\end{equation}
The positive mining recall (PMR) measures each time the fraction of TPs in the final Top-$k$ selected as the positive set.
The class mining recall (CMR) measures the fraction of distinct TPs selected from a class in one training epoch.
Table~\ref{table:capability} shows that as PMR and mining R@1 increase with more stages, higher fine-tuning performance on action recognition is expected.
However, PMR seems to serve as a better performance indicator for being in proportion to improvement.

Furthermore, we provide a breakdown of full PMR during the entire pretraining process for 100 epochs on UCF101 in Figure~\ref{fig:full_PMR}. 
The results show CPR gradually increases its PMR from 34.0\% to 40.3\%. 
On the other hand, the PMR of baseline CoCLR is sluggish between 35.2\% and 37.4\%. 
It can be found that CPR indeed can mine more true positives by leveraging both RGB and flow views while baseline CoCLR suffers from false positives for mining only in a single view. 
This is the crucial factor to support why CPR has better performance than baseline CoCLR. 
In summary, PMR seems to serve as a better performance indicator for being in proportion to improvement.

\begin{table*}[]
\centering
\begin{tabular}{ll|ccc|rr}
\hline
Method & \multicolumn{1}{l|}{Year} & \multicolumn{1}{l}{Dataset} & \multicolumn{1}{l}{Resolution} & \multicolumn{1}{l|}{Architecture} &  \multicolumn{1}{l}{UCF101} & \multicolumn{1}{l}{HMDB51} \\ \hline
VCOP~\cite{Xu2019-uv} & 2019 & UCF101 & 16 × $112^2$ & R(2+1)D-26 & 72.4 & 30.9 \\
VCP~\cite{Luo2020-gu} & 2020 & UCF101 & 16 × $112^2$ & C3D & 68.5 & 32.5 \\
IIC~\cite{Tao2020-tb} & 2020 & UCF101 & 16 × $112^2$ & R3D-18 & 74.4 & 38.3 \\
PacePred~\cite{Wang2020-hs} & 2020 & UCF101 & 16 × $112^2$ & R(2+1)D & 75.9 & 35.9 \\
PRP~\cite{Yao2020-gv} & 2020 & UCF101 & 16 × $112^2$ & C3D & 69.1 & 34.5 \\
TT~\cite{Jenni2020-wc} & 2020 & UCF101 & 16 × $112^2$ & R3D-18 & 77.3 & 47.5 \\
CoCLR-RGB~\cite{Han2020-od} & 2020 & UCF101 & 32 × $128^2$ & S3D & 81.4 & 52.1 \\
DSM~\cite{Wang2021-id} & 2021 & UCF101 & 16 × $112^2$ & C3D & 70.3 & 40.5 \\
STS~\cite{Wang2021-na} & 2021 & UCF101 & 16 × $112^2$ & R3D-18 & 77.8 & 40.7 \\
CMD~\cite{Huang2021-ds} & 2021 & UCF101 & 16 × $112^2$ & R3D-26 & 76.6 & 47.2 \\
MFO~\cite{Qian2021-lm} & 2021 & UCF101 & 32 × $128^2$ & S3D & 74.3 & 37.2 \\
$\rm Vi^{2}CLR$~\cite{diba2021vi2clr} & 2021 & UCF101 & 32 × $128^2$ & S3D & 82.8 & 52.9 \\ 
MCN~\cite{Lin2021-md} & 2021 & UCF101 & 32 × $128^2$ & S3D & 82.9 & 53.8 \\
\hline
CoCLR-Flow(+CPR) &  & UCF101 & 32 × $128^2$ & S3D & \textbf{83.8} & \textbf{54.8} \\ \hline
\end{tabular}
\caption{Comparison with SOTA action recognition on UCF101 and HDMB51 based on pretraining on UCF101}
\label{table:recog}
\end{table*}

To further quantify the mining quality across classes, we count the number of distinct TPs selected for each action class.
Figure~\ref{fig:bar_pos} illustrates the statistics from 10 randomly chosen classes with 18 instances each in the last training epoch.
CPR succeeds in selecting all the distinct TPs from both \textit{Surfing} and \textit{UnevenBars} classes while discovering much less from the \textit{Hammering} class.
Through visual inspection, \textit{Hammering} is difficult with motions at different camera angles in varied scenes.
In contrast, \textit{Swing} and \textit{Surfing} are easy to mine for having regular motion patterns and consistent background.
Furthermore, we list the CMRs of the Top-3 and the bottom 3 classes respectively. 
The Top-3 classes are \textit{Diving}(100\%) , \textit{PommelHorse}(100\%), and \textit{UnevenBars}(100\%).
On the other hand, the bottom 3 classes are \textit{Lunges}(27.8\%), \textit{Haircut}(33.3\%) and \textit{HandstandWalking}(33.3\%). We demonstrate video frames from these classes above to visualize their content including human action and background. In Figure~\ref{fig:top3_bottom3}, sampled frames in the bottom 3 classes cover different camera angles and varied backgrounds, which increases the difficulty of mining full video instances in each class.  In contrast, it is simple to discover entire video instances in each Top-3 class because these classes represent a relatively consistent background and standard motion with a fixed pattern.
To sum up, out of all the UCF101 classes, CPR scores higher CMR than baseline CoCLR in 48 classes while the baseline mines better only in 20 classes. 
It is even in the rest classes.
Overall, CPR achieves the median CMR of \textbf{83.3\%} across all the classes, which is \textbf{5.5\%} improvement over the baseline CoCLR with the median CMR of 77.8\%.

Besides quantitative measurements, we visualize the Top-5 positive examples mined from the baseline CoCLR and CPR for qualitative comparison.
In Figure~\ref{fig:exp_pos_visual}, the baseline mining heavily relies on the motions from optical flows and tends to select false positives (FPs) with similar motion patterns to the query. 
Instead, our CPR alternately mines from both RGB and flow views to discover positive examples with similar appearances and motions to the query. 
Even the only FP still contains visually similar motions and context as the query. 
This indicates that CPR is able to effectively filter out potential FPs from a single view. 
Consequently, CPR facilitates learning representations from more diverse TPs compared with the baseline mining.

\noindent \textbf{Applicability.} 
In addition to CoCLR, CPR is also applied to IIC~\cite{Tao2020-tb} in the ablation of progressive training, demonstrating the general applicability.
Particularly, IIC uses memory banks instead of momentum encoders to maintain features as well as frame difference residuals as motion views.
It focuses on generating hard negatives from the query video by repeating or shuffling the frames but there is no positive example mining.
With CPR, IIC gains significant performance improvement in both downstream tasks on UCF101 in Table~\ref{table:trans} where CoCLR is also listed to show consistent performance boost.
This suggests that CPR is generally applicable whether or not the existing approach has positive example mining in mind.

\subsection{Comparison with State-of-the-arts}
\label{sec:sota}

As CPR aims to benefit existing work in terms of better positive example mining, our focus is to show how much improvement an existing work can be enhanced with CPR to compete with newer SOTAs.
CoCLR~\cite{Han2020-od} is chosen as it already has positive example mining in mind.

\noindent \textbf{Video Retrieval.} To validate the effectiveness of learned representations with CPR, we evaluate the nearest neighbor video retrieval on both UCF101 and HMDB51. 
Specifically, the top-$k$ video retrieval recalls for $k=1, 5, 10, 20$ are computed as the performance metrics.
As shown in Table~\ref{table:retrieval}, CoCLR-Flow(+CPR) outperforms the the other SOTA methods in all recall metrics on both datasets.
We achieve the best top-1 recall of \textbf{56.7\%} on UCF101 and \textbf{24.4\%} on HMDB51, outperforming the the latest SOTA MCN~\cite{Lin2021-md} by up to 2.9\% based on the same backbone. 
Moreover, CPR also gains much more improvement at higher top-$k$ metrics. 
Since video retrieval does not require fine-tuning and leaves little room for manipulation, positive example mining from diverse positive examples across distinct videos is likely the key to learning effective representations.

\noindent \textbf{Action Recognition.} In table~\ref{table:recog}, we compare our method with SOTAs on video action recognition. All methods are applied in a fully finetuning setting that finetunes all layers on the downstream task. Pretrained on UCF101, CoCLR-Flow(+CPR) outperforms all the previous SOTAs fine-tuned on UCF101 and HMDB51 with accuracies of \textbf{83.8\%} and \textbf{54.8\%} based on the same or comparable backbone and resolutions as illustrated in Table~\ref{table:recog}. 

\section{Conclusion}
\label{sec:conclusion}

In this work, we propose the Cascade Positive Retrieval (CPR) for self-supervised video representation learning and extensively explore the design space of positive example mining configurations.
We find that more mining stages in the cascade likely improves the performance.
The positive selection ratio on the contrary works better if set to a smaller number.
The progressive training with an incremental final Top-$k$ selection could bring potential improvement.   
Beyond the R@1 mining retrieval recall by CoCLR\cite{Han2020-od}, we further measure the mining quality quantitatively in PMR and CMR that seem to correlate with downstream task performance better. 
Moreover, the mining quality is also visualized for qualitative comparison.
Finally, we evaluate the transfer performance from UCF101 to UCF101 and HMDB51 that is either SOTA or competitive in both video retrieval and action recognition.
Aside from promising results, our CPR can be applied to existing work easily regardless of a specific SSL framework used or not.
Nonetheless, the gap from the supervised contrastive performance upper bound remains, suggesting the necessity of follow-up research for even better mining in self-supervised representation learning. 
In the future, we plan to facilitate the application of CPR to existing work, automate the hyperparameter search for improved mining quality, and examine the scalability of transfer learning from large-scale dataset.

\clearpage
{\small
\bibliographystyle{ieee_fullname}
\bibliography{egbib}
}

\clearpage
\appendix
\section*{Appendices}

\begin{figure*}[h]
  \begin{center}
    \includegraphics[width=\textwidth]{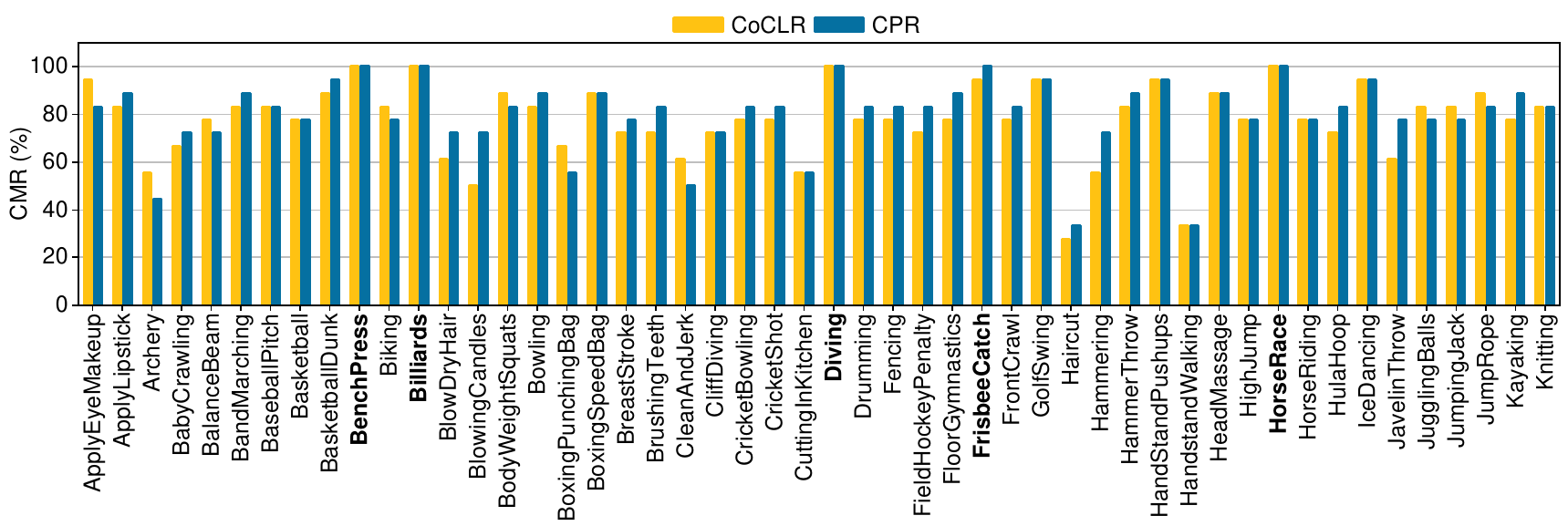}
  \end{center}

  \begin{center}
    \includegraphics[width=\textwidth]{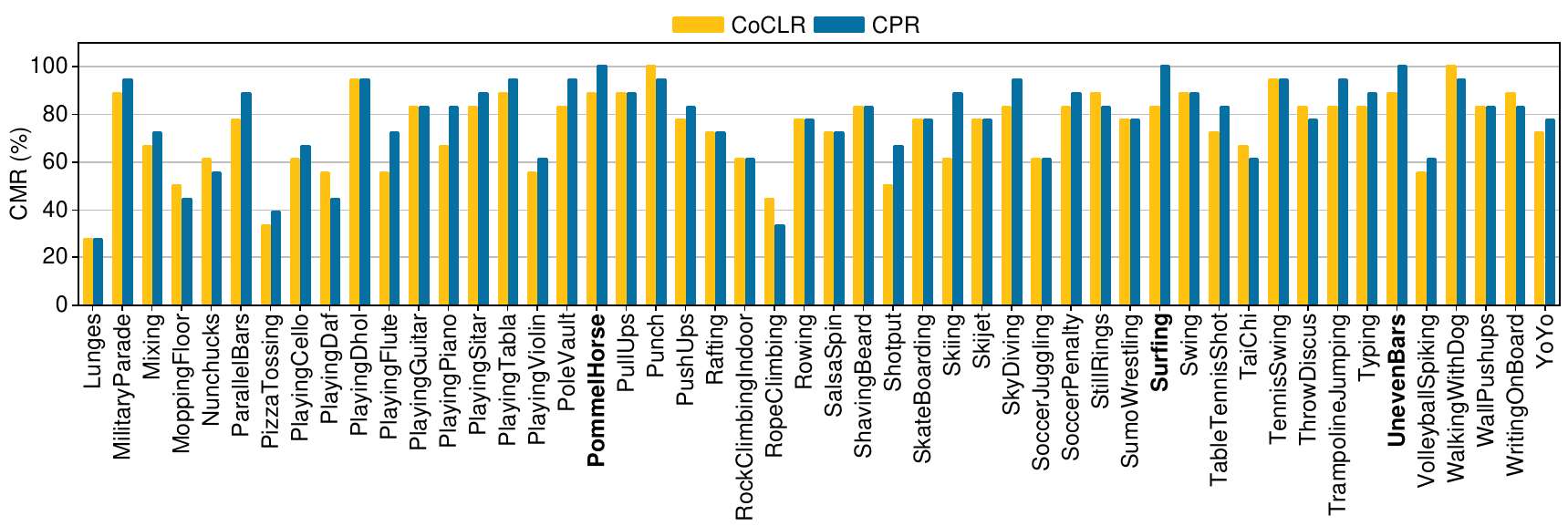}
  \end{center}
  \caption{Class Mining Recall (CMR) per action class on UCF101. There are 101 action classes which are listed in alphabetical order. The upper bar chart covers the first 50 action classes while the lower bar chart covers the rest of the action classes. Eight action classes appeared in bold font represent cases where CPR achieves 100\% CMR. They are 1. \textit{BenchPress}, 2. \textit{Billiards}, 3. \textit{Diving}, 4. \textit{HorseRace}, 5. \textit{FrisbeeCatch}, 6. \textit{PommelHorse}, 7. \textit{UnevenBars}, and 8. \textit{Surfing}.
  Note that we measure the CMR of both approaches in their last training epoch.}
  \label{fig:CMR_bar_chart}
\end{figure*}

\section{Implementation Details}
\label{sec:details}
\subsection{Self-supervised Pretraining}
\label{details:pretraining}
We use MoCo~\cite{He2019-wo} as the contrastive learning framework and S3D~\cite{Xie2018-fr} as the feature extractor to implement CPR. Note that MoCo is not required but is useful to save memory usage. Therefore, it is simply a coincidence that the baseline CoCLR~\cite{Han2020-od} uses MoCo and we have applied CPR to other work not using MoCo such as IIC as well.
At pretraining stage, two fully connected layers (FC1024→ReLU→FC128) are used as a projection head after the global average pooling layer to obtain the embedding features but the projection head is removed for the model to perform downstream tasks. 
Following CoCLR we set the momentum to 0.999, the temperature to 0.07, and the size of the queue to 2048 on every dataset Training each model on UCF 101 , we use ADAM as our optimizer with an initial learning rate of $10^{-3}$ and weight decay of $10^{-5}$, where the learning rate is multiplied by 0.1 at 300 and 350 epochs.

\subsection{Action Recognition}
\label{details:action}
For action recognition task, we use ADAM to optimize the model for 500 epochs with a batch size of 16 on two GPUs.
The initial learning rate is set to $10^{-3}$, where the learning rate is decayed by 0.1 at 400 and 450 epoch respectively. 
The momentum is 0.9 and the weight decay is $10^{-3}$.
At evaluation stage, we follow the practice of CoCLR to uniformly sample 32 frames from each video, perform ten-crop to 128×128 pixels, and then average their predictions to become the final video prediction.

\section{Additional Results}
\label{sec:addtional_results}

\subsection{Class Mining Recall (CMR) }
\label{details:action}
To evaluate the overall mining quality, we further define Class Mining Recall (CMR) in Eq~\ref{eq:cls_recall} to measure how a model is able to successfully mine distinct true positives from a certain class in one training epoch. 
\begin{equation}
\label{eq:cls_recall}
\footnotesize
Class~Mining~Recall=\frac{\#Distinct~TP~Selected}{\#Total~Class~Instances}
\end{equation}
As shown in Figure~\ref{fig:CMR_bar_chart}, we present the full CMR in the UCF101 classes. Closer inspection of the figure reveals that CPR has the Top-3 classes are \textit{Diving} (100\%) , \textit{PommelHorse} (100\%), and \textit{UnevenBars} (100\%).
On the other hand, the bottom 3 classes are \textit{Lunges} (27.8\%), \textit{Haircut} (33.3\%), and \textit{HandstandWalking} (33.3\%).  Furthermore, we demonstrate video frames from these classes above to visualize their content including human action and background. 


In the overall evaluations of all 101 classes, CPR scores higher CMR than baseline CoCLR in 48 classes while the baseline mines better only in 20 classes. It is even in the rest classes. Regarding the classes that the baseline has higher CMR, it may that those positive classes highly correlate with a single view while CPR sometimes does not help much after visual inspection. A further comparison of the number of 100\% CMR between both methods, CPR obtains eight perfect CMRs across 101 classes, which exceeds two classes compared to the baseline getting six. In addition, we evaluate the entire performance of CPR by applying median CMR. Our approach achieves the median CMR of \textbf{83.3\%} across all the classes, which outperforms the baseline with the median CMR of 77.8\%. This outcome shows a great improvement with a margin of \textbf{5.5\%}. In summary, these empirical evidence validates the effectiveness of our approach in mining the higher quality positives than the baseline. 
\vspace{15.5cm} \\

\end{document}